\documentclass[conference]{IEEEtran}
\IEEEoverridecommandlockouts
\usepackage{cite}
\usepackage{amsmath,amssymb,amsfonts}
\usepackage{algorithmic}
\usepackage{graphicx}
\usepackage{textcomp}
\usepackage{xcolor}
\def\BibTeX{{\rm B\kern-.05em{\sc i\kern-.025em b}\kern-.08em
    T\kern-.1667em\lower.7ex\hbox{E}\kern-.125emX}}

\usepackage{hyperref}
\ifCLASSOPTIONcompsoc
    \usepackage[caption=false,font=normalsize,labelfont=sf,textfont=sf{subfig}
\else
    \usepackage[caption=false,font=footnotesize]{subfig}
\fi

\newtheorem{definition}{\textbf{Definition}}
\newtheorem{lemma}{\textbf{Lemma}}
\newtheorem{theorem}{\textbf{Theorem}}

\begin{document}

\title{The Transferability of Downsampled Sparse Graph Convolutional Networks}

\author{
\IEEEauthorblockN{
Qinji Shu\IEEEauthorrefmark{1}, 
Hang Sheng\IEEEauthorrefmark{1}, 
Feng Ji\IEEEauthorrefmark{3},
Hui Feng\IEEEauthorrefmark{1}\IEEEauthorrefmark{2}, 
Bo Hu\IEEEauthorrefmark{1}\IEEEauthorrefmark{2}
}

\IEEEauthorblockA{
\IEEEauthorrefmark{1}
School of Information Science and Technology, Fudan University, Shanghai 200433, China}

\IEEEauthorblockA{
\IEEEauthorrefmark{2}
State Key Laboratory of Integrated Chips and Systems, Fudan University, Shanghai 200433, China}

\IEEEauthorblockA{
\IEEEauthorrefmark{3}
 School of Electrical and Electronic Engineering, Nanyang Technological University, Singapore
\\
Email: \{qjshu22, hsheng20\}@m.fudan.edu.cn, jifeng@ntu.edu.sg, \{hfeng, bohu\}@fudan.edu.cn
}
}

\maketitle

\begin{abstract}
To accelerate the training of graph convolutional networks (GCNs) on real-world large-scale sparse graphs, downsampling methods are commonly employed as a preprocessing step. However, the effects of graph sparsity and topological structure on the transferability of downsampling methods have not been rigorously analyzed or theoretically guaranteed, particularly when the topological structure is affected by graph sparsity. In this paper, we introduce a novel downsampling method based on a sparse random graph model and derive an expected upper bound for the transfer error. Our findings show that smaller original graph sizes, higher expected average degrees, and increased sampling rates contribute to reducing this upper bound. Experimental results validate the theoretical predictions. By incorporating both sparsity and topological similarity into the model, this study establishes an upper bound on the transfer error for downsampling in the training of large-scale sparse graphs and provides insight into the influence of topological structure on transfer performance.
\end{abstract}


\section{Introduction}
Graph convolutional networks (GCNs) for large-scale graphs have gained significant attention due to their broad range of applications in domains that involve graph-structured data \cite{yin2023deepdrug},\cite{ishiai2023graph}. Despite their success across various tasks, training GCNs on large-scale graphs poses substantial challenges, primarily because the computation involved demands considerable storage and time resources.

In order to accelerate the training of large-scale graphs, downsampling to smaller-scale graphs is a prevalent preprocessing technique \cite{liu2021sampling},\cite{zhang2023survey}. The conventional approach involves sampling similar smaller subgraphs from the original large-scale graph, training the GCN on these subgraphs, and then applying the trained parameters to the original large-scale graph \cite{yao2021blocking,chen2018fastgcn,chiang2019cluster}. The effectiveness of transfer depends on the similarity between different objects, which in this context refers to the similarity in topological structures. This transfer method raises a critical question: how do different topologies affect transferability?

While large-scale GCNs are gaining increasing attention, real-world graphs often exhibit significant sparsity. The neighborhoods of most nodes are limited in size and do not expand proportionally with the overall graph size \cite{strogatz2001exploring}. For example, in a small community, everyone may know each other, whereas in a larger community, the number of acquaintances for a single person remains limited \cite{ricci2022thinned}. Consequently, the connection patterns of graph topologies vary across sparse graphs of different sizes, which in turn affects the effectiveness of downsampling methods based on topological similarity.

Related research on the transferability of GCNs either neglects the impact of topological structure, or focuses exclusively on graphs generated by sparse models without considering \textbf{downsampling on a specific sparse graph}. From a theoretical perspective, some studies \cite{cong2020minimal}, \cite{li2022generalization} have examined how specific sampling methods affect transferability, but have not considered how the initial topological properties of sparse graphs influence transferability. Other works \cite{keriven2020convergence} in the domains of GCNs and random graphs have investigated the transferability of GCNs influenced by sparse topological structures. However, their findings are limited to graphs with an increasing average degree, and crucially, they do not consider the effects of downsampling, particularly for sparse graphs where the average degree remains bounded. To address bounded-degree sparse graphs, Le and Jegelka \cite{le2024limits} adopt a perspective of sampling GCNs directly from graph operator limits, without considering the underlying graph topological properties.

This paper bridges this gap by linking the transferability of downsampling in sparse graphs to the topological structure of the graphs.
To better represent the sparse graph data structures commonly observed in real-world scenarios, we propose a simpler sparse random graph model, based on \cite{borgs2018sparse} and \cite{ji2023sparse}, that allows for adjustable levels of sparsity, such as generating large-scale sparse graphs with a fixed expected average degree. Building on the work of \cite{ruiz2021graphon} and \cite{ruiz2023transferability}, we also introduce a method for large-scale graph downsampling that preserves similar topological structures. 
As the main result, we derive a theorem regarding the transferablity errors of GCNs, which indicates that large-scale graphs with smaller sizes and higher expected average node degrees exhibit improved transferability. Additionally, increasing the sampling rate can further enhance transfer performance.

\textbf{Contributions:}
\begin{itemize}
\item We propose a downsampling method based on a sparse graph model, and establish a connection between the model's sparsity and the topological similarity of the method.
\item We prove a transferability theorem that bounds a distance between the GCN outputs of the original large-scale sparse graph and its downsampled smaller-scale graphs. This bound is related to the original graph sizes, the expected average node degree, and the downsampling rates.
\end{itemize}

\section{Preliminaries}

In this section, we will introduce graph convolutional networks and demonstrate how each of their layers can be expressed in the form of a graph signal convolution. Furthermore, to compare the outputs of GCNs of different dimensions and scales, we transform the GCN output into a piece-wise function over a certain interval through interpolation.

Graph convolutional networks are based on graph convolution operator, which is defined through the graph adjacency matrix and graph signal $(\mathbf{S}_n,\boldsymbol{x}_n)$: the adjacency matrix $\mathbf{S}_n$ is a graph shift operator of graph $\mathbf{G}_n$, based on it, the convolution operator is defined as \cite{segarra2017optimal}:
\begin{equation}
\begin{aligned}
    h(\mathbf{S}_n) * \boldsymbol{x}_n 
    &= \sum^{K-1}_{k=0} h_k \mathbf{S}_n^k \boldsymbol{x}_n
    &= h({\mathbf{S}_n}) \boldsymbol{x}_n,
\end{aligned}
\end{equation}
the weights $\{h_k\}, k \in \{0,...,K-1\}$ are the graph filter taps, the number of weights is only related to the highest order $(K-1)$ of convolution operator, and not influenced by the scale $n$ of graphs. 

Let $\Phi(\mathbf{S}_n, \boldsymbol{x}_n, \boldsymbol{\mathcal{H}})$ denote a graph convolutional network, dealing with the graph and graph signal $(\mathbf{S}_n, \boldsymbol{x}_n)$. And $\boldsymbol{\mathcal{H}}$ denotes the weights of all layers, i.e., for a GCN with $L$ layers and its $l$th layer outputs $F_l$ features for each node, we have $\boldsymbol{\mathcal{H}}(l) \in \mathbb{R}^{F_{l-1} \times F_l}.$ And the input $\boldsymbol{x}_n$ of GCN $\Phi(\mathbf{S}_n, \boldsymbol{x}_n, \boldsymbol{\mathcal{H}})$ has $F_0 = 1$ feature.

In layer $l \in \{1,2,...,L\}$, the number of features is denoted by $F_l \in \{F_1,F_2,...,F_L\}$. And the output graph signals can be set in matrix form $\mathbf{X}_n^l \in \mathbb{R}^{n \times F_l}$, the input signal matrix is $\mathbf{X}_n^{l-1} \in \mathbb{R}^{n \times F_{l-1}}$. 

For one layer $l$, the aggregation and propagation mechanism are given by:
\begin{equation}
\begin{aligned}
    \mathbf{X}_n^l &= \sigma \left( \hat{\mathbf{S}}_{{n}}  {\mathbf{X}}_{{n}} ^{l-1} \boldsymbol{\mathcal{H}}(l) \right),
\end{aligned}
\end{equation}
where $\hat{\mathbf{S}}_n = {h}({\mathbf{S}}_n) $ is a determined polynomial of ${\mathbf{S}}_n$.

For each feature vector of layer $l$, $\boldsymbol{x}_{f_l}$ is the $f_l$the column of $\mathbf{X}_n^l$, denotes the $f_l$th feature \cite{ruiz2023transferability}:
\begin{equation}
\begin{aligned}
    \boldsymbol{x}_{f_l} = \sigma \left( \sum_{f_{l-1}=1}^{F_{l-1}} {{h}}_{f_{l-1},f_{l}}(\mathbf{S}_n) * \boldsymbol{x}_{f_{l-1}} \right),
\end{aligned}
\end{equation}
where $\{ {{h}}_{f_{l-1},f_{l}} = h_{f_{l-1},f_{l}}  {{h}}(\mathbf{S}_n) \}$ are convolution filters for input signal $\boldsymbol{x}_{f_{l-1}}$. We can see $\boldsymbol{x}_{f_l}$ is the sum of $F_{l-1}$ convolution's results.

In order to compare GCN outputs of different graph sizes, we convert the outputs of vertor form into induced continuous form:
\begin{equation}
\overline{ \Phi} (\mathbf{S}_n, \boldsymbol{x}_n, \boldsymbol{\mathcal{H}}) = \xi^{\mathfrak{s}} \left( \Phi(\mathbf{S}_n, \boldsymbol{x}_n, \boldsymbol{\mathcal{H}}) \right),
\end{equation}
where $\xi^{\mathfrak{s}}(\cdot)$ is the piece-wise interpolation function, details in our extended version\cite{shu2024transferability}.

\section{A Sparse Graph Model and Downsampling}
To capture topological similarity across GCNs with different sizes, many studies \cite{keriven2020convergence}, \cite{ruiz2023transferability}, \cite{maskey2023transferability}, \cite{neuman2023transferability} leverage a class of symmetric and measurable functions knowned as graphons. Graphons can be viewed as the limit objects of sequences of graphs with similar topologies, which is a powerful fundamental model to generate graphs with such structures. However, they typically generate dense or isolated graphs \cite{lovasz2012large}.

In the following section, based on graphons we introduce a sparse random graph model that allows for the setting of different sparsity levels, as well as a downsampling method for large-scale sparse graphs to maintain topological similarity. 

We also highlight the fundamental differences between these two methods. Sparsity implies that the probability of connections between nodes decreases as the graph size increases \cite{orbanz2014bayesian}, while the similarity in topology indicates that the connection probability maintains an unchanging pattern \cite{ruiz2021graphon}.

\subsection{A Sparse Random Graph Model}
Graph sparsity refers to the growth trend of the number of edges in a graph as the graph's size increases \cite{orbanz2014bayesian}: 
\begin{definition}
    Let $\{\mathbf{G}_n\}$ be a sequence of graphs, where $\mathbf{G}_n$ has $n$ nodes and $e_n$ edges. Graphs of the sequence $\{\mathbf{G}_n\}$ are called sparse if, as $n$ increases, $e_n$ is of size $O(n)$. Graphs are called dense if $e_n$ is of size $\Omega(n^2)$.
\end{definition}

Accordingly, sparsity can be represented by the edge density, which denotes the proportion of existing edges out of all $\binom{n}{2}=\frac{n(n-1)}{2}$ possible edges \cite{diestel2024graph}. In this paper, we focus primarily on the expected value of this edge density:
\begin{equation}
    {\epsilon}(n):=\frac{\mathbb{E}\{e_n\}}{n(n-1)/2}.
\end{equation}
For sparse graphs, their edge density is of size $O(n^{-1})$; for dense graphs, their edge density is of size $\Omega(1)$:
\begin{equation}
    \begin{aligned}
        \mathrm{Sparse:} \quad {\epsilon}(n)=\frac{O(n)}{n(n-1)/2}=O(n^{-1})
        \\
        \mathrm{Dense:} \quad {\epsilon}(n)=\frac{\Omega(n^2)}{n(n-1)/2}=\Omega(1).
    \end{aligned}
\end{equation}
Edge density is a commonly used measure of sparsity in networks and graphs \cite{goswami2018sparsity}, \cite{ravazzi2021learning}. By using edge density instead of the total number of edges, we can assess the sparsity of a graph from the perspective of the overall node connection probability. As the graph size grows, the overall node connection probability remains constant or even increases for dense graphs, whereas it continuously decreases for sparse graphs.

\textbf{A random graph model with adjustable sparsity}. 
The sparse random graph model $(W_{\mathbb{R}_+}, t_n, X)$ involves a kernel $W_{\mathbb{R}_+}$, a scale function $t_n$, and a signal function $X$. 
The kernel $W_{\mathbb{R}_+}$ is a symmetric function: $\mathbb{R}_+^2 \to [0,1]$, and $W_{\mathbb{R}_+} \in L^{1}({\mathbb{R}_+}^2)$.  The scale function $t_n$ is an increasing function: $\mathbb{Z}_+ \to \mathbb{R}_+$. The signal function $X$ is defined on $[0,1]$: $[0,1] \to \mathbb{R}$.

To generate a sparse random graph $\mathbf{G}_n$, we first sample a graphon $W_{t_n}$ from the kernel $W_{\mathbb{R}_+}$. The scale function $t_n$ limits the truncation range of the kernel $W_{\mathbb{R}_+}$, and we obtain graphon $W_{t_n}$ through scaling the limited kernel:
\begin{equation}
    W_{t_n} (u, v) =  W_{\mathbb{R}_+} (u t_n, v t_n),
\end{equation}
where $u, v\in [0,1]$ are latent node features. The sequence of graphons generated by the sparse model converges to the kernel $W_{\mathbb{R}_+}$ in the stretched cut distance as $N$ increases \cite{ji2023sparse}.

Then we sample the random graph $\mathbf{G}_n$ and its graph signal $\boldsymbol{x}_n$ from the graphon $W_{t_n}$ and the signal function $X$. $n$ points $\{u_1,u_2,...,u_n\}$ are sampled independently and uniformly at random from $[0,1]$, as latent vertex features:
\begin{equation}
u_i \stackrel{iid}{\sim} \mathrm{unif}(0,1)\    \mathrm{for}\ 1 \leq i \leq n,
\end{equation}
and the vertex feature $\boldsymbol{x}_n(i)$ is obtained from $X$, the edge connection probability $p(i,j)$ is obtained from $W$, based on which the connection of edges is determined by Bernoulli distributions:
\begin{equation}
\begin{aligned}
&\mathbf{S}_n(i,j) \sim \mathrm{Ber}(W_{t_n}(u_i,u_j))\  \mathrm{for}\ 1 \leq i,j \leq n,
\\
&\boldsymbol{x}_n(i) = X(u_i)\  \mathrm{for}\ 1 \leq i \leq n.
\end{aligned}
\end{equation}

The graphon $W_{t_N}$ can be decomposed into eigenvalue and eigenfunctions as follows \cite{ruiz2021graphon}: $W_{t_n}(u,v) = \sum_{i \in \mathbb{Z} \setminus \{0\}} \lambda_i \phi_i(u) \phi_i(v)$, where $\{\lambda_i\} \in [-1,1]$ and $\{\phi_i\}: [0,1] \to \mathbb{R}$. Since the graphon sequence $\{W_{t_n}\}$ convergences to a limit in the stretched cut distance \cite{borgs2018sparse}, we analyze the graphon $W_{t_n}$ and signal function $X$ in their stretched forms: $W^{\mathfrak{s}}_{t_n} (u,v)=W_{t_n} (\sqrt{\Vert W_{t_N} \Vert_1}u, \sqrt{\Vert W_{t_N} \Vert_1}v)$ and $X^{\mathfrak{s}}(u) = X(\sqrt{\Vert W_{t_N}\Vert_1}u)$. The stretched graphon $W^{\mathfrak{s}}_{t_n}$ has corresponding eigenvalues and eigenfunctions given by: $\lambda^{\mathfrak{s}}_i = \lambda_i / \sqrt{\Vert W_{t_N} \Vert_1}$ and $\phi^{\mathfrak{s}}_i(u) = \phi_i(\sqrt{\Vert W_{t_N} \Vert_1}u)$ \cite{ji2023sparse}.

\textbf{Spasity of the sparse random model.} Since the edge probabilities $\{p_{ij}\}$ are obtained from $W_{t_n}$ through uniform distributions, the expected number of edges is given by:
\begin{equation}
\begin{aligned}
    \mathbb{E} \{e_n\} &= \mathbb{E}_p \mathbb{E}_e \left\{ \sum_{i=1}^{n} \sum_{j<i} e_{ij} \right\} 
    = \mathbb{E}_p \left\{ \sum_{i=1}^{n} \sum_{j<i} p_{ij} \right\} 
    \\
    &= \frac{n(n-1)}{2} \iint_{[0,1]^2} W_{t_n} dudv,
\end{aligned}
\end{equation}
where $\mathbb{E}_p$ is the expectation about $\{p_{ij}\}$ based on uniform distributions, and $\mathbb{E}_e$ is the expectation about $\{e_{ij}\}$ based on Bernoulli distributions. Therefore, the edge density of the sparse graph model is a function of graph size $n$:
\begin{equation}
    \epsilon(n) = \iint_{[0,1]^2} W_{t_n} dudv = \frac{ \iint_{[0,t_n]^2} W_{\mathbb{R}_+} dudv } {t_n^2}.
\end{equation}
After selecting the kernel $W_{\mathbb{R}_+}$, we adjust the scale function $t_n$ to adjust the sparsity of graphs, e.g., when $W_{\mathbb{R}_+} \in L_1(\mathbb{R}_+^2)$, and $t_n = \sqrt{n}$, we have $\epsilon(n) \sim \Theta(\frac{1}{n})$, and the expected average degree $\mathbb{E}\{ \bar{d}(n) \}$ is:
\begin{equation}
    d(n) := \mathbb{E}\{ \bar{d}(n) \} = (n-1)\epsilon(n) \approx n\epsilon(n),
\end{equation}
where $\bar{d}(n)$ is the average degree. So the expected average degree $d(n)$ is $\Theta(1)$, denoted by a patameter $d$.

The graphs generated by a sparse graph model have the same level of sparsity across different scales, reflected as $\epsilon(n)$ or $d(n)$, but their topological structures are not identical, reflected as $W_{t_n}$. A sparse random graph model controls the number of edges that increase as the graph size grows, ensuring that the expected edge density $\epsilon(n)$ is a decreasing function of $n$. 
From the perspective of edge connection probability,  graphs of different scales under the same sparse random graph model have different structures because they are sampled from varying graphons $W_{t_n}$.

\subsection{A Downsampling Method for Large-scale Sparse Graphs}
The downsampling method for large-scale sparse graphs differs from the method used to generate sparse graphs. Sparse graph models generate graphs of different scales, each corresponding to a distinct graphon obtained through sampling, in order to maintain the sparsity across different graph sizes. In contrast, the downsampling of a sparse graph typically targets a specific large-scale sparse graph, aiming to preserve the same topological structure during the downsampling process.

We obtain a large-scale sparse graph and nodes features $(\mathbf{S}_N, \boldsymbol{x}_N)$ from the sparse graph model:
\begin{equation}
\begin{aligned}
&\mathbf{S}_N(i,j) \sim \mathrm{Ber}(W_{t_N}(u_i,u_j))\  \mathrm{for}\ 1 \leq i,j \leq N,
\\
&\boldsymbol{x}_N(i) = X(u_i)\  \mathrm{for}\ 1 \leq i \leq N.
\label{downsample_large}
\end{aligned}
\end{equation}
And a smaller-scale graph and nodes features $(\mathbf{S}_n, \boldsymbol{x}_n)$ are derived through downsampling:
\begin{equation}
\begin{aligned}
&\mathbf{S}_n(i,j) \sim \mathrm{Ber}(W_{t_N}(u_i,u_j))\  \mathrm{for}\ 1 \leq i,j \leq n \ (n \ll N),
\\
&\boldsymbol{x}_n(i) = X(u_i)\  \mathrm{for}\ 1 \leq i \leq n.
\label{downsample_small}
\end{aligned}
\end{equation}

Since the downsampling goal is to maintain the same topology, the smaller-scale graph and the large-scale graph share the same graphon function $W_{t_N}$. Therefore, the essence of downsampling a large-scale sparse graph is to obtain a smaller graph with a similar structure, rather than one with identical sparsity. As a result, the downsampling process does not involve sampling a new graphon $W_{t_n}$, as it would in the sparse graph model.

By combining the sparse graph model with the downsampling method, we obtain the initial large-scale graph with sparsity, and we preserve the invariant topological structure at a specific scale through downsampling.

\section{Transferability of Graph Convolutional Networks}
Considering the downsampling process of the large-scale graph mentioned above, we obtained a smaller-scale graph by downsampling from a large-scale sparse graph while maintaining a similar topological structure. Generally, the transferability of downsampling training is influenced both by the sampling rate and by the sparsity and scale of the large-scale graph. Therefore, we aim to understand the impact of varying the sampling rate on transferability for an original graph with fixed sparsity and scale, as well as the effect of different sparsity levels and scales of the original graph on transferability when the sampling rate is fixed.

\begin{theorem}
    (GCN Downsampling Transferability). 
    Let $(\mathbf{S}_N, \boldsymbol{x}_N)$ denote the large-scale sparse graph and graph signal obtained from the sparse random model with $t_N = \sqrt{N}$ and $\epsilon(N) \sim \Theta(\frac{1}{N})$, let $(\mathbf{S}_n, \boldsymbol{x}_n)$ be the smaller-scale graph and graph signal sampled by the large-scale graph downsampling method. Consider the $L$-layer GCNs $\Phi(\widetilde{\mathbf{S}}_N, \boldsymbol{x}_N, \boldsymbol{\mathcal{H}})$ and $\Phi(\widetilde{\mathbf{S}}_n, \boldsymbol{x}_n, \boldsymbol{\mathcal{H}})$, where $F_0 = F_L = 1$ and $F_l = F$ for $1 \leq l \leq L-1$. Then, under Assumptions 1-4 \cite{shu2024transferability} it holds
    \begin{equation}
        \begin{aligned}
        \mathbb{E} &\left\{ \left\Vert \xi^{\mathfrak{s}} \left( \Phi(\widetilde{\mathbf{S}}_N, \boldsymbol{x}_N, \boldsymbol{\mathcal{H}}) \right) - \xi^{\mathfrak{s}} \left( \Phi(\widetilde{\mathbf{S}}_n, \boldsymbol{x}_n, \boldsymbol{\mathcal{H}}) \right) \right\Vert_2 \right\} 
        \\
        \leq 
        &C_h  \left\{ 1 +  C_{\mathbb{R}_+} \sqrt{\frac{N}{d}} \left( 1 + \sqrt{ \frac{N}{n} } \right) \right\}
        + C_m
        \\
        &+ C_s { \left( \frac{N}{d} \right)}^{\frac{1}{4}} \left( \frac{1}{\sqrt{N}} + \frac{1}{\sqrt{n}} \right), 
        \end{aligned}
    \end{equation}
    where $\mathbb{E}\{\cdot\}$ is about the uniform distributions and Bernoulli distributions in the downsampling process (\ref{downsample_large}), (\ref{downsample_small}).$\widetilde{\mathbf{S}}_n =  {\mathbf{S}_n} / (n \sqrt{2\epsilon(n)} )$, $d$ is the expected average degree of the large-scale sparse graph, and $C_h = \sqrt{2} LF^{L-1} ||X^{\mathfrak{s}}||_2 A_h$, $C_{\mathbb{R}_+} =  A_{R_+} / \sqrt{6} $, $C_s = A_s / \sqrt{6\sqrt{2}}$, $C_m = 4LF^{L-1} ||X||_2 \Delta {h}(\lambda^{\mathfrak{s}})$ are deterministic parameters about GCN and the sparse graph model. In $C_m$, $\Delta h(\lambda^{\mathfrak{s}}) = \mathop{\min}_{k \in \mathbb{R}} \mathop{\max}_{ \lambda^{\mathfrak{s}}_i}\left\{ |{h}(\lambda^{\mathfrak{s}}_i) - k| \right\}$ for all convolutional filters ${h}$, and $\{\lambda^{\mathfrak{s}}_i, i \in \mathbb{Z} \setminus \{0\} \}$ are eigenvalues of the stretched graphon $W^{\mathfrak{s}}_{t_N}$.
\label{theorem_of_downsampling}
\end{theorem}

\textbf{Discussion}. We derive conclusions applicable to sparse graph models with \textbf{varying sparsity levels}, not limited to cases $t_N = \sqrt{N}$ and $\epsilon(N)\sim \Theta(\frac{1}{N})$. These conclusions align with the trends concerning $N$, $n$, and $d$ observed in the results of the aforementioned model. For further details, please refer to our full version \cite{shu2024transferability}, which includes all related proofs.

From the first term of the inequality in Theorem 1, we observe that a higher sampling rate $N/n$ can reduce the upper bound of the transfer error, thereby enhancing transfer performance. In contrast, a larger original graph scale $N$ and a smaller expected average degree $d$ of the original large-scale graph $\mathbf{G}_n$ tend to increase the upper bound, resulting in poorer transfer performance. 

The second term of the inequality highlights the impact of the frequency response of the graph convolutional network on transfer error: the smoother the frequency response curve $h(\lambda)$, the smaller this term becomes, e.g., when the frequency response is constant, this term is reduced to zero.

The third term of the inequality indicates that, the error related to sampling node features, decreases with larger graph sizes $N$ and $n$. This implies that a larger downsampled graph will have a smaller transfer error. Although a larger original graph size $N$ leads to smaller errors in the third term, the increase in error from the first term is more significant, resulting in an overall trend of increasing error.

\section{Experiments}
In the following, we conduct the large-scale graph downsampling method sampled from a sparse random graph model. We consider untrained GCNs with initially random weights to focus on the transferability error between the original large-scale sparse graph and the downsampled smaller-scale graphs, instead of learning some specific tasks.

We consider the following kernel:
\begin{equation}
    W_{\mathbb{R}_+}(u,v) = \left\{
    \begin{aligned}
        &e^{-u}e^{-v} &\quad &u \neq v \\
        &0 &\quad &u=v. 
    \end{aligned}
    \right
    .
\end{equation}
Based on it, we adjust the edge density by the following kernel form:
\begin{equation}
    W_{\mathbb{R}_+}^{'}(u,v) = \left\{
    \begin{aligned}
        &c_d W_{\mathbb{R}_+}(u,v) &\quad &c_dW_{\mathbb{R}_+}(u,v) \leq 1 \\
        &1 &\quad &c_dW_{\mathbb{R}_+}(u,v) > 1, 
    \end{aligned}
    \right
    .
\end{equation}
where we increase $c_d$ to increase the average degree expectation ${d}$.
And we set the scale function to be $t_n = \sqrt{n}$.



\begin{figure}[!t]
    \centering
    \subfloat[]{\includegraphics[width=0.48\columnwidth]{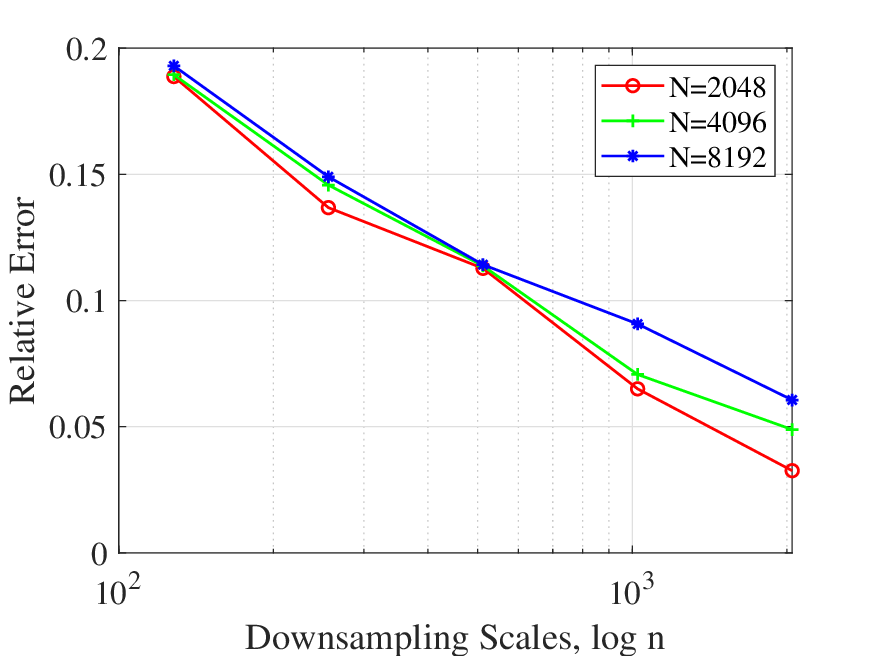}
    \label{scales}}
    \subfloat[]{\includegraphics[width=0.48\columnwidth]{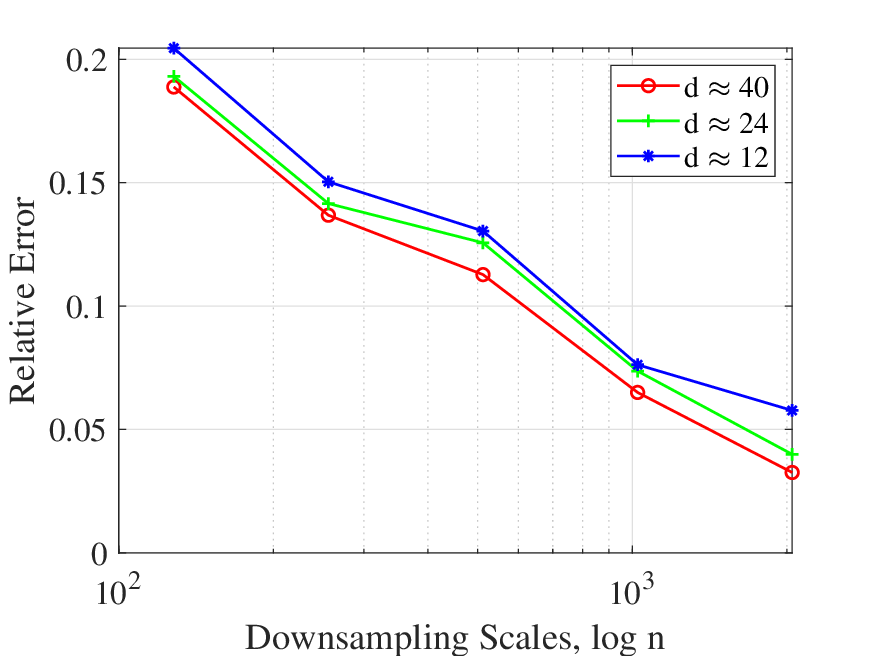}
    \label{degrees}}
    \caption{Experiments about transferability error (a) with different original scales, (b) with different expected average degrees. In both sets of experiments, we also examine the effect of varying the sampling rate, specifically the impact of sampling larger smaller-scale graphs.}
    \label{fig:enter-label}
\end{figure}

We use the relative errors to evaluate the transferability: 
\begin{equation}
    e_r = \frac{||\xi \left( \Phi(\widetilde{\mathbf{S}}_N, \boldsymbol{x}_N, \boldsymbol{\mathcal{H}}) \right) - \xi \left( \Phi(\widetilde{\mathbf{S}}_n, \boldsymbol{x}_n, \boldsymbol{\mathcal{H}}) \right) ||_{L_2} } {|| \xi \left( \Phi(\widetilde{\mathbf{S}}_N, \boldsymbol{x}_N, \boldsymbol{\mathcal{H}}) \right) ||_{L_2}}.
\end{equation}

The first part of experiments is about how the original graph scales $N$ influence GCNs transferability. We adjust $c_d$ to set the average degree expectation: ${d} \approx 40$. We set three groups of different original sizes $\{2048, 4096, 8192 \}$, and each group samples smaller-scale graphs of different sizes $\{128, 256, 512, 1024, 2048\}$. From the experiment results in \figurename \ref{scales}, the transferability errors decrease as the sampling sizes $n$ increase, and the group of larger original sizes $N$ tend to have bigger errors.

The second part of experiments is about how the average degree expectation influences GCNs transferability. We set the original graph sizes to be $2048$, and sampling sizes to be $\{128, 256, 512, 1024, 2048\}$. We adjust $c_d$ to set different average degrees $\{40, 24, 12 \}$. From the experiment results in \figurename \ref{degrees}, the transferability errors decrease as the sampling sizes $n$ increase, and the group of larger degree expectation tend to have smaller errors.

Our experimental results validate the theoretical findings, demonstrating that smaller original graph scales, higher expected average degrees, and increased sampling rates contribute to improved transferability performance.

\newpage
\appendices

\section{Preliminaries}

\subsection{Graphon Signal Processing}
\textbf{Graphon and Graphon Signal}. A Graphon $W$ is a symmetric function: $[0,1]^2 \to [0,1]$. 
A Graphon signal $X$ is a function: $[0,1] \to \mathbb{R}$.

Let ($\mathbf{G}_n$, $\boldsymbol{x}_n$) denotes a graph and graph signal sampled from graphon and graphon signal ($W$, $X$) at random with $n$ nodes ($n \in \mathbb{Z}^+$), and its adjacency matrix is denoted by $\mathbf{S}_n$.

To generate a random graph $\mathbf{G}_n$ and its graph signal ${\boldsymbol{x}}_n$, $n$ points $\{u_1,u_2,...,u_n\}$ are sampled independently and uniformly at random from $[0,1]$, as latent vertex features \cite{ruiz2021graphon}:
\begin{equation}
u_i \stackrel{iid}{\sim} \mathrm{unif}(0,1)\    \mathrm{for}\ 1 \leq i \leq n,
\nonumber
\end{equation}
and the vertex feature $\boldsymbol{x}_n(i)$ is obtained from $X$, the edge connection probability $p(i,j)$ is obtained from $W$, based on which the connection of edges is determined by Bernoulli distributions:
\begin{equation}
\begin{aligned}
&\mathbf{S}_n(i,j) \sim \mathrm{Ber}(W(u_i,u_j))\  \mathrm{for}\ 1 \leq i,j \leq n,
\\
&\boldsymbol{x}_n(i) = X(u_i)\  \mathrm{for}\ 1 \leq i \leq n.
\nonumber
\end{aligned}
\end{equation}

Graphon convoluion operator is based on $(W,X)$, similarly the graphon shift operator is firstly introduced \cite{ruiz2021graphon}:
\begin{equation}
    (T_{W}X)(v) = \int_0^1 W(u,v) X(u) du.
\end{equation}
Baed on it, the graphon convolution operator is defined as:
\begin{equation}
\begin{aligned}
{h}(W) * X &= \sum^{K-1}_{k=0} h_k (T_{W}X)^{(k)},
\\
(T_W X)^{(k)}(v) &= \int_0^1 W(u,v) (T_W X)^{(k-1)}(u) du.
\end{aligned}
\end{equation}
The graphon $W$ can also be decomposed into eigenvalues $\{\lambda_i\} \in [-1,1]$ and eigenfunctions $\{\phi_i\}: [0,1] \to \mathbb{R}$, $i \in \mathbb{Z} \setminus \{0\}$:
\begin{equation}
    W(u,v) = \sum_{i \in \mathbb{Z} \setminus \{0\}} \lambda_i \phi_i(u) \phi_i(v).
\end{equation}

The eigenfunctions are normalized and orthogonal. Let $\hat{X}_i = \int_0^1 \phi_i(u) X(u) du$ denotes the projection of graphon signal $X$ onto eigenfunction $\phi_i, i \in \mathbb{Z} \setminus \{0\}$, then we have:
\begin{equation}
\begin{aligned}
    (T_{W}X)(v) 
    &= \int_0^1 W(u,v) X(u) du \\
    &= \int_0^1 \sum_{i \in \mathbb{Z} \setminus \{0\}} \lambda_i \phi_i(u) \phi_i(v) X(u) du \\
    &= \sum_{i \in \mathbb{Z} \setminus \{0\}} \lambda_i \phi_i(v) \hat{X}_i,
    \\
    (T_W X)^{(k)}(v) 
    &= \int_0^1 W(u,v) (T_W X)^{(k-1)}(u) du \\
    &= \sum_{i \in \mathbb{Z} \setminus \{0\}} \lambda_i^k \phi_i(v) \hat{X}_i.
\end{aligned}
\end{equation}
Therefore, the graphon convolution operator can be translated into a filter form:
\begin{equation}
\begin{aligned}
    {h}(W) * X 
    &= \sum^{K-1}_{k=0} h_k (T_{W}X)^{(k)} \\
    &= \sum^{K-1}_{k=0} h_k \sum_{i \in \mathbb{Z} \setminus \{0\}} \lambda_i^k \phi_i(v) \hat{X}_i \\
    &= \sum_{i \in \mathbb{Z} \setminus \{0\}} \left( \sum^{K-1}_{k=0} h_k \lambda_i^k \right) \phi_i(v) \hat{X}_i \\
    &= \sum_{i \in \mathbb{Z} \setminus \{0\}} h(\lambda_i) \phi_i(v) \hat{X}_i.
\end{aligned}
\end{equation}
Then we can see the frequency response $h(\lambda_i)$, which is a polynomial with parameters $\{h_k\}$.

\subsection{Graphon Convolutional Networks}

(Graphon Convolution Networks \cite{ruiz2020graphon}). 
Let $\Phi(W, X, \boldsymbol{\mathcal{H}})$ denote a graphon convolution network, dealing with the graphon and graphon signal $(W, X)$. And $\boldsymbol{\mathcal{H}}$ denotes the weights of all layers, i.e., for a WNN with $L$ layers and its $l$th layer outputs $F_l$ features $\{ X_{f_l} \}, f_l \in \{1,2,...,F_l\}$, we have $\boldsymbol{\mathcal{H}}(l) \in \mathbb{R}^{F_{l-1} \times F_l}.$ And the input $X$ of WNN $\Phi(W, X, \boldsymbol{\mathcal{H}})$ has $F_0 = 1$ feature.

Similarly, for each feature function of layer $l$, the aggregation and propagation mechanism are given by:
\begin{equation}
    X_{f_l} = \sigma \left( \sum_{f_{l-1}=1}^{F_{l-1}} {h}_{f_l,f_{l-1}} (W) * X_{f_{l-1}} \right).
\end{equation}
\\
After L layers' convolutions, the output is:
\begin{equation}
    X_L = \Phi(W, X, \boldsymbol{\mathcal{H}}), \    \mathrm{when}\ F_L = 1.
\end{equation}

\subsection{The Continuous Form of Graphs and Signals}
\begin{definition}
(Induced Graphon and Graphon Signal \cite{ruiz2021graphon} \cite{ruiz2023transferability}). Let $(\overline{W}_{n},\overline{X}_{n})$ denote the graphon and graphon signal induced by graph and graph signal $(\mathbf{S}_n,\boldsymbol{x}_n)$.
\end{definition}

To obtain $(\overline{W}_{n},\overline{X}_{n})$, the equal spaced partition $\{I_1,I_2,...,I_n\}$ of $[0,1]$ is constructed, here $I_i = [\frac{i-1}{n},\frac{i}{n}) $ for $i \in \{1,2,...,n-1\}$ and $I_n = [\frac{n-1}{n},1]$. Then $(\overline{W}_{n},\overline{X}_{n})$ are obtained as
\begin{equation}
\begin{aligned}
    \overline{W}_{n} (u,v) &= \sum_{i=1}^n \sum_{j=1}^n \mathbf{S}_n (i,j) \times \mathbb{I}(u \in I_i) \mathbb{I}(v \in I_j),
    \\
    \overline{X}_{n} (u) &= \sum_{i=1}^n \boldsymbol{x}_n (i) \times \mathbb{I}(u \in I_i),
\end{aligned}
\end{equation}
where $\mathbb{I}$ is the indicator function.

We consider the induced graphon and graphon signal $(\overline{W}_{n},\overline{X}_{n})$ as a continuous form of the graph and graph signal $(\mathbf{S}_n,\boldsymbol{x}_n)$, and the above conversion method can be referred to as:
\begin{equation}
\begin{aligned}
    &\overline{W}_{n} (u,v) = \xi ( \mathbf{S}_n ),
    \\
    &\overline{X}_{n} (u) = \xi ( \boldsymbol{x}_n ).
\end{aligned}
\end{equation}

As the graph sequence converges in the stretched cut distance \cite{borgs2018sparse}, we use the stretched graphon and graphon signal as another continuous forms, based on induced graphon and graphon signal.

(Stretched Graphon and Graphon Signal \cite{ji2023sparse}). The stretched graphon $\overline{W}^{\mathfrak{s}}_n$ and signal $\overline{X}^{\mathfrak{s}}_n$ are defined as:
\begin{equation}
\begin{aligned}
    &{W}^{\mathfrak{s}}_n(u,v) = \overline{W}_n(\Vert \overline{W}_n \Vert^{1/2}_1 u , \Vert \overline{W}_n \Vert^{1/2}_1 v),
    \\
    &{X}^{\mathfrak{s}}_n(u) = \overline{X}_n(\Vert \overline{W}_n \Vert^{1/2}_1 u).
\end{aligned}
\end{equation}
To facilitate the comparison of GCN outputs at different scales, it is desirable for the transformed continuous form to be within the same interval. Therefore, we use the expected value instead of $\Vert \overline{W}_n \Vert_1$:
\begin{equation}
    \mathbb{E} (\Vert \overline{W}_n \Vert_1) = \frac{\mathbb{E}(2e_n)}{n^2} = 2\epsilon(n). 
\end{equation}
\begin{definition}
    (Stretched Continuous Forms)We defined the stretched continuous forms $(\overline{W}^{\mathfrak{s}}_n(u,v), \overline{X}^{\mathfrak{s}}_n(u))$ of graph and graph signal $(\mathbf{S}_n,\boldsymbol{x}_n)$ as:
\begin{equation}
    \begin{aligned}
        &\xi^{\mathfrak{s}}(\mathbf{S}_n) = \overline{W}^{\mathfrak{s}}_n(u,v) = \overline{W}_n(\sqrt{2\epsilon(n)} u , \sqrt{2\epsilon(n)} v),
        \\
        &\xi^{\mathfrak{s}}(\boldsymbol{x}_n) = \overline{X}^{\mathfrak{s}}_n(u) = \overline{X}_n(\sqrt{2\epsilon(n)} u).
    \end{aligned}
\end{equation}
\end{definition}

Similar to graphon signal processing and WNNs, the signal processing and convolutional networks can be defined based on integral approaches, with the only difference being the variation in the integration interval. We denote the stretched graphon convolutional networks (SWNNs) as:
\begin{equation}
    \Phi(\xi^{\mathfrak{s}} ( \mathbf{S}_n ), \xi^{\mathfrak{s}} ( \boldsymbol{x}_n ), \boldsymbol{\mathcal{H}}).
\end{equation}

\subsection{The Connection Between GCNs and SWNNs}
To compare transferability across different graph scales, the first method applies a GCN to the graph and its signals to generate outputs, which are then transformed into their continuous forms for comparison. The second method converts the graph and its signals into continuous forms and then uses a SWNN to generate continuous outputs for comparison. By adjusting the adjacency matrix, both methods can produce identical continuous results.
\begin{lemma}
Let $(\mathbf{S}_n, \boldsymbol{x}_n)$ denote the graph and node features to be processed by GCN and SWNN respectively, which have the same activation function $\sigma$, layers and weights $\boldsymbol{\mathcal{H}}$, then we have:
\begin{equation}
\xi^{\mathfrak{s}} \left( \Phi(\mathbf{S}_n / (n\sqrt{2\epsilon(n)}, \boldsymbol{x}_n, \boldsymbol{\mathcal{H}}) \right)
= \Phi \left( \xi^{\mathfrak{s}} ( \mathbf{S}_n ), \xi^{\mathfrak{s}} ( \boldsymbol{x}_n ), \boldsymbol{\mathcal{H}} \right),
\end{equation}
for GCN we use the adjacency matrix $\mathbf{S}_n/(n\sqrt{2\epsilon(n)})$, and SWNN's input graphon is $\xi^{\mathfrak{s}} ( \mathbf{S}_n )$, then the continuous outputs of GCN and SWNN are the same. 
\label{G_W}
\end{lemma}

\textit{Proof of Lemma 1:} Let $\overline{W}^{\mathfrak{s}}_n$ denotes $\xi^{\mathfrak{s}} ( \mathbf{S}_n )$, let $\overline{X}^{\mathfrak{s}}_n$ denotes $\xi^{\mathfrak{s}} ( \boldsymbol{x}_n )$, for the shift operator we have:
\begin{equation}
\begin{aligned}
    T_{\overline{W}^{\mathfrak{s}}_n}\overline{X}^{\mathfrak{s}}_n &= \int_0^{(2\epsilon(n))^{-\frac{1}{2}}} \overline{W}^{\mathfrak{s}}_n(u,v) \overline{X}^{\mathfrak{s}}_n(u) du
    \\
    &=\sum_{i=1}^n \frac{1}{n\sqrt{2\epsilon(n)}} \left( \sum_{j=1}^n \mathbf{S}_n (i,j) \times  \mathbb{I}(v \in I_j) \right) \boldsymbol{x}_n(i) 
    \\
    &= \xi^{\mathfrak{s}} \left( \frac{1}{n\sqrt{2\epsilon(n)}} \mathbf{S}_n \boldsymbol{x}_n \right),
\end{aligned}
\end{equation}
after iteration, for multi-times shift operator we have:
\begin{equation}
    (T_{\overline{W}^{\mathfrak{s}}_n}\overline{X}^{\mathfrak{s}}_n)^{(k)} = \xi^{\mathfrak{s}} \left( \left(\frac{ \mathbf{S}_n }{n\sqrt{2\epsilon(n)}} \right)^k  \boldsymbol{x}_n \right),
\end{equation}
therefore, for the convolutional operators sharing same weights:
\begin{equation}
    {h}(\overline{W}^{\mathfrak{s}}_n) * \overline{X}^{\mathfrak{s}}_n = \xi \left( h \left(\frac{\mathbf{S}_n}{n\sqrt{2\epsilon(n)}} \right) * \boldsymbol{x}_n \right).
\end{equation}
The aggregation of GCNs and SWNNs layers can be represented as convolutional operations. When GCN and WNN share the same weights and their initial inputs satisfy the conditions: $\overline{W}^{\mathfrak{s}}_n = \xi^{\mathfrak{s}} ( \mathbf{S}_n )$ and $\overline{X}^{\mathfrak{s}}_n = \xi^{\mathfrak{s}} ( \boldsymbol{x}_n$ ), it holds:
\begin{equation}
\xi^{\mathfrak{s}} \left( \Phi(\mathbf{S}_n / (n \sqrt{2\epsilon(n)}, \boldsymbol{x}_n, \boldsymbol{\mathcal{H}}) \right)
= \Phi(\xi^{\mathfrak{s}} ( \mathbf{S}_n ), \xi^{\mathfrak{s}} ( \boldsymbol{x}_n ), \boldsymbol{\mathcal{H}}).
\end{equation}

\section{Proof of Sampling Lemmas}
To generate a sparse random graph $\mathbf{G}_N$, we first sample a graphon $W_{t_N}$ from the kernel $W_{\mathbb{R}_+}$ through a scale function $t_N$. Then we sample the sparse large-scale graph $(\mathbf{S}_N, \boldsymbol{x}_N)$ from graphon $W_{t_N}$ and the signal function $X$. In order to make potential node features not differentiated by the arbitrary order, we sort the potential features by numerical values $\{u_1 \leq u_2 \leq ... \leq u_N \}$. Each edge $e_{ij}$ of $\mathbf{G}_N$ is related to a connection probability $p_{ij}$, and we denote all these probabilities as a matrix $\mathbf{P}_N$. We transform $\mathbf{S}_N, \boldsymbol{x}_N, \mathbf{P}_N$ into their continuous form:
\begin{equation}
    \begin{aligned}
        \overline{W}^{\mathfrak{s}}_N &= \xi^{\mathfrak{s}} ( \mathbf{S}_N ),
        \\
        \overline{P}^{\mathfrak{s}}_N &= \xi^{\mathfrak{s}} ( \mathbf{P}_N ),
        \\
        \overline{X}^{\mathfrak{s}}_N &= \xi^{\mathfrak{s}} ( \boldsymbol{x}_N ).
    \end{aligned}
\end{equation}

We also consider the following assumptions:

\textbf{AS1} : 
The kernel $W_{\mathbb{R}_+}$ of the sparse model is $A_{\mathbb{R}_+}$-Lipschitz, i.e. $|W_{\mathbb{R}_+}(u_1,v_1) - W_{\mathbb{R}_+}(u_2,v_2)| \leq A_{\mathbb{R}_+}(|u_1-u_2|+|v_1-v_2|)$.

\textbf{AS2} \cite{ruiz2023transferability} : 
The signal function $X$ of the sparse model is $A_{s}$-Lipschitz, i.e. $|X(u_1) - X(u_2)| \leq A_s |u_1 - u_2|$.

\textbf{AS3} \cite{ruiz2023transferability} : 
The convolutional filters ${h}$ are $A_{h}$-Lipschitz and non-amplifying, i.e. $|h(\lambda)| \leq 1$.

\textbf{AS4} \cite{ruiz2023transferability} : 
The activation functions $\sigma(\cdot)$ are normalized-Lipschitz, i.e. $|\sigma(x_1) - \sigma(x_2)| \leq |x_1 - x_2|$, and $\sigma(0)=0$.

The assumptions AS1, AS2 requires the kernel $W_{\mathbb{R}_+}$ and the signal function $X$ to be continuous functions with bounded slopes. These assumptions are primarily concerned with the Lipschitz property, without imposing additional restrictions, such as memberships in $L^p$-space. Assumption AS3, related to convolutional filters, is straightforward, as their responses to the spectrum are essentially polynomial functions, which are naturally Lipschitz continuous. Assumption AS4, concerning activation functions, applies to commomly used activation functions, such as ReLU and Tanh.

\begin{lemma}
    Let $X$ be a $A_{s}$-Lipschitz signal function of the sparse random graph model, and let $\overline{X}_N$ and $\overline{X}^{\mathfrak{s}}_N$ be the two continuous forms of the graph signal $\boldsymbol{x}_N$ obtained from $X$. The $L_2\ norm$ of $\overline{X}_N - X$ and $\overline{X}^{\mathfrak{s}}_N - X^{\mathfrak{s}}$, where $X^{\mathfrak{s}} = \xi^{\mathfrak{s}} (\boldsymbol{x}_n)$, satisfies:
    \begin{equation}
    \begin{aligned}
        &\mathbb{E} \left\{ \left\Vert \overline{X}_N - X \right\Vert_2 \right\} \leq \frac{A_s}{\sqrt{6N}},
        \\
        &\mathbb{E} \left\{ \left\Vert \overline{X}^{\mathfrak{s}}_N - X^{\mathfrak{s}} \right\Vert_2 \right\} \leq \frac{A_s}{\sqrt{6N 
\sqrt{2\epsilon(n) }}}.
    \end{aligned}
    \end{equation}
\label{lemma_s}
\end{lemma}
\textit{Proof.} As $\overline{X}_N$ is divided into equal spaced partition $\{I_1,I_2,...,I_N\}$ of $[0,1]$, here $I_i = [\frac{i-1}{N},\frac{i}{N}) $ for $i \in \{1,2,...,N-1\}$ and $I_N = [\frac{N-1}{N},1]$, $\overline{X}_N - X$ can also be divided into the equal partition. Using the assumption about the signal function $|X(u_1) - X(u_2)| \leq A_s |u_1 - u_2|$, we get:
\begin{equation}
\begin{aligned}
    \left\Vert \overline{X}_N - X \right\Vert_2 &= \sqrt{ \sum_{i=1}^N \int_{I_i} ( X(u_i) - X(u) )^2 du }
    \\
    &\leq A_s \sqrt{\sum_{i=1}^N \int_{I_i} (u_i - u)^2 du}.
\end{aligned}
\end{equation}
For the right side of the above inequality, we calculate the integral and sum the results:
\begin{equation}
\begin{aligned}
    &A_s \sqrt{\sum_{i=1}^N \int_{I_i} (u_i - u)^2 du} 
    \\
    &= A_s \sqrt{ \sum_{i=1}^N \frac{i^3-(i-1)^3}{3N^3} - \frac{2i-1}{N^2} u_i + \frac{1}{N} u_i^2 }
    \\
    &= A_s \sqrt{\frac{1}{3} - \frac{1}{N^2} \sum_{i=1}^N (2i-1)u_i + \frac{1}{N} \sum_{i=1}^N u_i^2 }.
\end{aligned}
\end{equation}
Let's consider the expectation of the difference:
\begin{equation}
\begin{aligned}
    &\mathbb{E} \left\{ || \overline{X}_N - X ||_2 \right\} 
    \\
    &\leq \sqrt{ \mathbb{E} \left\{ \sum_{i=1}^N \int_{I_i} ( X(u_i) - X(u) )^2 du \right\} }
    \\
    &\leq A_s \sqrt{ \mathbb{E} \left\{ \frac{1}{3} - \frac{1}{N^2} \sum_{i=1}^N (2i-1)u_i + \frac{1}{N} \sum_{i=1}^N u_i^2 \right\} }.
\label{E_s}
\end{aligned}
\end{equation}
According to the order statistic \cite{okoyo2016order}, the probability density function of $u_i$ is $f_{u_i} = \mathrm{Beta}(i, n-i+1)$, and we have:
\begin{equation}
\begin{aligned}
    \mathbb{E}(u_i) &= \frac{i}{N+1},
    \\
    \mathbb{E}(u_i^2) &= \frac{i^2+i}{(N+1)(N+2)},
\label{E_u}
\end{aligned}
\end{equation}
and substitute (\ref{E_u}) into (\ref{E_s}), through calculating the sum of the series we get:
\begin{equation}
    \mathbb{E} \left\{ || \overline{X}_N - X ||_2 \right\} \leq \frac{A_s}{\sqrt{6N}}.
\end{equation}
Because $\overline{X}^{\mathfrak{s}}_N - X^{\mathfrak{s}}$ is devived from the extension of the domain of $\overline{X}_N - X$, we get:
\begin{equation}
    \mathbb{E} \left\{ \left\Vert \overline{X}^{\mathfrak{s}}_N - X^{\mathfrak{s}} \right\Vert_2 \right\} \leq \frac{A_s}{\sqrt{6N \sqrt{2\epsilon(n) }}}.
\end{equation}

\begin{lemma}
    Let $W_{\mathbb{R}_+}$ be an $A_{\mathbb{R}_+}$-Lipschitz kernel of the sparse random graph model, let $W_{t_N}$ be the graphon sampled from $W_{\mathbb{R}_+}$ through the scale function $t_N$, and let $\overline{W}_N$ and $\overline{W}^{\mathfrak{s}}_N$ be the continuous form of the sparse graph $\mathbf{S}_N$ obtained from $W_{t_N}$. The $L_2\ norm$ of $\overline{W}_N - W_{t_N}$ satisfies:
    \begin{equation}
        \mathbb{E} \left\{ \left\Vert \overline{W}_N - W_{t_N} \right\Vert_{2} \right\} \leq \frac{\sqrt{L_1 - L_2^2}}{t_N} + \frac{2A_{\mathbb{R}_+}t_N}{\sqrt{6N}},
    \end{equation}
    The $L_2\ norm$ of $\overline{W}^{\mathfrak{s}}_N - W^{\mathfrak{s}}_{t_N}$ satisfies:
    \begin{equation}
        \mathbb{E} \left\{ \left\Vert \overline{W}^{\mathfrak{s}}_N - W^{\mathfrak{s}}_{t_N} \right\Vert_{2} \right\} \leq \frac{\sqrt{L_1 - L_2^2}}{t_N {(2\epsilon(n))}^{\frac{1}{4}} } + \frac{2A_{\mathbb{R}_+}t_N}{\sqrt{6N \sqrt{2\epsilon(n)} }},
    \end{equation}
\label{l_w}
\end{lemma}
where $L_1 = ||W_{\mathbb{R}_+}||_{L_2}$, $L_2 = ||W_{\mathbb{R}_+}||_{L_2}$ in $[0,t_N]^2$.
\\
\textit{Proof.} Using the triangle inequality, we can write the norm difference as:
\begin{equation}
\begin{aligned}
    ||\overline{W}_N - W_{t_N}||_{2} 
    &= ||\overline{W}_N - \overline{P}_N + \overline{P}_N - W_{t_N}||_{2} 
    \\
    &\leq ||\overline{W}_N - \overline{P}_N||_2 + ||\overline{P}_N - W_{t_N}||_{2}.
\label{W_tr}
\end{aligned}
\end{equation}

For the first part of the right side in (\ref{W_tr}), $\overline{W}_N$ stores edges, and $\overline{P}_N$ stores connection probabilities, we need to consider both the expectaion of latent features' uniform distribution $u_i \sim \mathrm{Uni}([0,1])$, and the expectation of edges' bernouli distribution $e_{ij} \sim \mathrm{Ber}(p_{ij})$:
\begin{equation}
    \mathbb{E} \left\{ ||\overline{W}_N - \overline{P}_N||_{2} \right\} = \mathbb{E}_U \mathbb{E}_B \left\{ ||\overline{W}_N - \overline{P}_N||_{2} \right\},
\end{equation}
and for the expectation of edges' bernouli distribution, we have:
\begin{equation}
    \mathbb{E}_B \left\{ (e_{ij} - p_{ij})^2 \right\} = p_{ij}(1 - p_{ij}),
\label{E_b}
\end{equation}
where $p_{ij} = W_{t_N}(u_i, u_j)$.

As two functions $\overline{W}_N$ and $\overline{P}_N$ are both piecewise interpolation function, $||\overline{W}_N - \overline{P}_N||_2$ can be divided into the equal partition:
\begin{equation}
\begin{aligned}
    ||\overline{W}_N - \overline{P}_N||_2 
    &= \sqrt{ \sum_{i=1}^N \sum_{j=1}^N \frac{(e_{ij} - p_{ij})^2}{N^2} }
    \\
    &= \sqrt{ \sum_{i=1}^N \sum_{j=1}^N \frac{(e_{ij} - W_{t_N}(u_i,u_j))^2}{N^2} },
\end{aligned}
\end{equation}
considering the expectation of above equation, and substituting (\ref{E_b}) we get:
\begin{equation}
\begin{aligned}
    \mathbb{E} &\left \{ ||\overline{W}_N - \overline{P}_N||_{2} \right\} 
    \\
    \leq &\sqrt{ \mathbb{E}_U  \mathbb{E}_B \left\{  \sum_{i=1}^N \sum_{j=1}^N \frac{(e_{ij} - W_{t_N}(u_i,u_j))^2}{N^2}  \right\} }
    \\
    &= \sqrt{ \mathbb{E}_U  \left\{  \sum_{i=1}^N \sum_{j=1}^N \frac{W_{t_N}(u_i,u_j) - W^2_{t_N}(u_i,u_j)}{N^2}  \right\} }.
\end{aligned}
\end{equation}
According to the Monte Carlo method \cite{hung2024review}, we have:
\begin{equation}
\begin{aligned}
    \mathbb{E}_U  \left\{  \sum_{i=1}^N \sum_{j=1}^N \frac{W_{t_N}(u_i,u_j)} {N^2}  \right\} &= \frac{L_1}{t_N^2}
    \\
    \mathbb{E}_U  \left\{  \sum_{i=1}^N \sum_{j=1}^N \frac{W^2_{t_N}(u_i,u_j)} {N^2}  \right\} &= \frac{L_2^2}{t_N^2},
\end{aligned}
\end{equation}
therefore, the expectation of $||\overline{W}_N - \overline{P}_N||_2$ holds
\begin{equation}
    \mathbb{E} \left \{ ||\overline{W}_N - \overline{P}_N||_{2} \right\} \leq \frac{\sqrt{L_1 - L_2^2}}{t_N}.
\label{w_l_1}
\end{equation}

For the second part of the right side in (\ref{W_tr}), we divide $||\overline{P}_N - W_{t_N}||_{2}$ into the equal partition:
\begin{equation}
\begin{aligned}
    ||\overline{P}_N - W_{t_N}||_{2} 
    = \sqrt{ \sum_{i=1}^N \sum_{j=1}^N \int_{I_i} \int_{I_j} \left( \overline{P}_N - W_{t_N} \right)^2 dudv },
\label{W_2}
\end{aligned}
\end{equation}
during the above equation, applying the $A_{\mathbb{R}_+}\ Lipschitz$, we get:
\begin{equation}
\begin{aligned}
    &\int_{I_i} \int_{I_j} \left( \overline{P}_N - W_{t_N} \right)^2 dudv
    \\
    &= \int_{I_i} \int_{I_j} \left( W(u_i, v_j) - W_{t_N}(u,v) \right)^2 dudv
    \\
    &\leq (A_{\mathbb{R}_+} t_N)^2 \int_{I_i} \int_{I_j} \left( |u_i - u| + |v_j - v| \right)^2 dudv
    \\
    &\leq 2(A_{\mathbb{R}_+} t_N)^2 \int_{I_i} \int_{I_j} \left( (u_i - u)^2 + (v_j - v)^2 \right) dudv.
    \label{W_A}
\end{aligned}
\end{equation}
Substituting (\ref{W_A}) into (\ref{W_2}), we repeat the proof process similar to lemma \ref{lemma_s} and we get:
\begin{equation}
    \mathbb{E} \left\{ ||\overline{P}_N - W_{t_N}||_{2} \right\} \leq \frac{2A_{\mathbb{R}_+}t_N}{\sqrt{6N}}.
\label{w_l_2}
\end{equation}
Combining (\ref{w_l_1}) and (\ref{w_l_2}), and considering the connection between $\overline{W}_N - W_{t_N}$ and $\overline{W}^{\mathfrak{s}}_N - W^{\mathfrak{s}}_{t_N}$ we prove lemma \ref{l_w}.

\section{Proof of Theorem \ref{theorem_con}}

\begin{lemma}
    Let $W_1$ and $W_2$ denote two graphos with eigenvalues given by $\{ \lambda_i(W_1) \}_{i \in \mathbb{Z}\setminus \{0\}}$ and $\{ \lambda_i(W_2) \}_{i \in \mathbb{Z}\setminus \{0\}}$, ordered according to their sign and in decreasing order of absolute value. Then, for all $i \in \mathbb{Z}\setminus \{0\}$, the following inequalities hold \cite{ruiz2023transferability}:
    \begin{equation}
        |\lambda_i(W_1) - \lambda_i(W_2)| \leq ||W_1 - W_2||_2.
    \end{equation}
\end{lemma}

Considering the stretched graphon $W^{\mathfrak{s}}_1$ and $W^{\mathfrak{s}}_2$ with the same stretch coefficient $\sqrt{2\epsilon(n)}$, the eigenvalues and eigenvectors are defined as \cite{ji2023sparse}: $\lambda^{\mathfrak{s}}_i = \lambda_i / \sqrt{2\epsilon(n)}$ and $\phi^{\mathfrak{s}}_i (u) = \phi( \sqrt{2\epsilon(n)} u )$, substituting into the above inequality, then we have:
\begin{equation}
    |\lambda^{\mathfrak{s}}_i(W^{\mathfrak{s}}_1) - \lambda^{\mathfrak{s}}_i(W^{\mathfrak{s}}_2)| \leq \frac{\Vert W^{\mathfrak{s}}_1 - W^{\mathfrak{s}}_2 \Vert_2}{(2\epsilon(n))^{\frac{1}{4}}}.
\end{equation}

\begin{theorem}
    Let $(\mathbf{S}_N, \boldsymbol{x}_N)$ be the large-scale graph and graph signal obtained from the sparse random graph model, let $(\mathbf{S}_n, \boldsymbol{x}_n)$ be the smaller-scale graph and graph signal  obtained from the downsampling method. For the graph convolutions $\boldsymbol{y}_N = {h}({\mathbf{S}}_N/(N\sqrt{2\epsilon(N)})) \boldsymbol{x}_N$ and $\boldsymbol{y}_n = {h}({\mathbf{S}}_n/(n\sqrt{2\epsilon(n)}) ) \boldsymbol{x}_n$, under assumptions 1 through 3 it holds:
    \begin{equation}
    \begin{aligned}
        &\mathbb{E}( ||\xi^{\mathfrak{s}} ( \boldsymbol{y}_N ) - \xi^{\mathfrak{s}} ( \boldsymbol{y}_n) ||_2 ) )
        \\
        &\leq 2A_h ||X||_2 \left( \frac{\sqrt{L_1 - L_2^2}}{t_N} + \frac{A_{\mathbb{R}_+}t_N}{\sqrt{6N}} + \frac{A_{\mathbb{R}_+}t_N}{\sqrt{6n}} \right)
        \\
        &\quad + \frac{A_s}{\sqrt{6}} \left( \frac{1}{\sqrt{N}} + \frac{1}{\sqrt{n}} \right) + 4\Delta h(\lambda) ||X||_2.
    \end{aligned}
    \end{equation}
    where $L_1 = ||W_{\mathbb{R}_+}||_{L_2}$, $L_2 = ||W_{\mathbb{R}_+}||_{L_2}$ in $[0,t_N]^2$ make $\sqrt{ 1 - \frac{L_2^2}{L_1} }$ decrease about $d$ and increase about $N$, and $\Delta h(\lambda) = \mathop{\min}_{k \in \mathbb{R}} \mathop{\max}_{ \lambda_i}\left\{ |{h}(\lambda_i ) - k| \right\}$ for all convolutional filters ${h}$, and $\{\lambda_i, i \in \mathbb{Z} \setminus \{0\} \}$ are eigenvalues of the graphon $W_{t_N}$.
\label{theorem_con}
\end{theorem} 

\textit{Proof of Theorem \ref{theorem_con}.} 
Let $(\overline{W}^{\mathfrak{s}}_N, \overline{X}^{\mathfrak{s}}_N)$ be the continuous forms of  the large-scale graph and graph signal $({\mathbf{G}}_N, \boldsymbol{x}_N)$ obtained from the sparse random graph model, let $(\overline{W}^{\mathfrak{s}}_n, \overline{X}^{\mathfrak{s}}_n)$ be the continuous forms of the smaller-scale graph and graph signal $({\mathbf{G}}_n, \boldsymbol{x}_n)$ obtained from the downsampling method.
From lemma \ref{G_W}, we know that
\begin{equation}
    ||\xi^{\mathfrak{s}} ( \boldsymbol{y}_N ) - \xi^{\mathfrak{s}} ( \boldsymbol{y}_n )||_2 = || T_{\overline{W}^{\mathfrak{s}}_N}  \overline{X}^{\mathfrak{s}}_N - T_{\overline{W}^{\mathfrak{s}}_n}  \overline{X}^{\mathfrak{s}}_n ||_2.
\end{equation}
Using the triangle inequality, we can write the above norm difference as:
\begin{equation}
\begin{aligned}
    &|| T_{\overline{W}^{\mathfrak{s}}_N} \overline{X}^{\mathfrak{s}}_N - T_{\overline{W}^{\mathfrak{s}}_n} \overline{X}^{\mathfrak{s}}_n ||_2
    \\
    &= || T_{\overline{W}^{\mathfrak{s}}_N} \overline{X}^{\mathfrak{s}}_N - T_{{W}^{\mathfrak{s}}_{t_N}} X^{\mathfrak{s}} + T_{{W}^{\mathfrak{s}}_{t_N}} X^{\mathfrak{s}} - T_{\overline{W}^{\mathfrak{s}}_n} \overline{X}^{\mathfrak{s}}_n ||_2
    \\
    &\leq || T_{\overline{W}^{\mathfrak{s}}_N} \overline{X}^{\mathfrak{s}}_N - T_{{W}^{\mathfrak{s}}_{t_N}} X^{\mathfrak{s}} ||_2 + || T_{{W}^{\mathfrak{s}}_{t_N}} X^{\mathfrak{s}} - T_{\overline{W}^{\mathfrak{s}}_n} \overline{X}^{\mathfrak{s}}_n ||_2.
\end{aligned}
\end{equation}
For the first part of the right side of the above inequality, we use the triangle inequality:
\begin{equation}
\begin{aligned}
    &|| T_{\overline{W}^{\mathfrak{s}}_N} \overline{X}^{\mathfrak{s}}_N - T_{{W}^{\mathfrak{s}}_{t_N}} X^{\mathfrak{s}} ||_2
    \\
    &= || T_{\overline{W}^{\mathfrak{s}}_N} \overline{X}^{\mathfrak{s}}_N - T_{\overline{W}^{\mathfrak{s}}_N} X^{\mathfrak{s}} + T_{\overline{W}^{\mathfrak{s}}_N} X^{\mathfrak{s}} - T_{{W}^{\mathfrak{s}}_{t_N}} X^{\mathfrak{s}} ||_2
    \\
    &\leq || T_{\overline{W}^{\mathfrak{s}}_N} \overline{X}^{\mathfrak{s}}_N - T_{\overline{W}^{\mathfrak{s}}_N} X^{\mathfrak{s}} ||_2 + ||T_{\overline{W}^{\mathfrak{s}}_N} X^{\mathfrak{s}} - T_{{W}^{\mathfrak{s}}_{t_N}} X^{\mathfrak{s}} ||_2,
\label{con_p_1}
\end{aligned}
\end{equation}
because the convolutional filters are non-amplifying, (\ref{con_p_1}) can be written as:
\begin{equation}
\begin{aligned}
    &|| T_{\overline{W}^{\mathfrak{s}}_N} \overline{X}^{\mathfrak{s}}_N - T_{{W}^{\mathfrak{s}}_{t_N}} X^{\mathfrak{s}} ||_2 
    \\
    &\leq ||\overline{X}^{\mathfrak{s}}_N - X^{\mathfrak{s}}||_2 + ||T_{\overline{W}^{\mathfrak{s}}_N} X^{\mathfrak{s}} - T_{{W}^{\mathfrak{s}}_{t_N}} X^{\mathfrak{s}} ||_2.
\end{aligned}
\end{equation}
Transforming to the frequency domain, and to simplify expression, we denote the eigenvalues and eigenfunctions of $\overline{W}^{\mathfrak{s}}_N$ and ${W}^{\mathfrak{s}}_{t_N}$ by $\{\overline{\lambda}^{\mathfrak{s}}_i, \overline{\phi}^{\mathfrak{s}}_i\}$ and $\{\lambda^{\mathfrak{s}}_i, \phi^{\mathfrak{s}}_i\}$, then we have:
\begin{equation}
\begin{aligned}
    &\left\Vert T_{\overline{W}^{\mathfrak{s}}_N} X^{\mathfrak{s}} - T_{{W}^{\mathfrak{s}}_{t_N}} X^{\mathfrak{s}} \right\Vert_2
    \\
    &= \bigg\Vert \sum_i h(\overline{\lambda}^{\mathfrak{s}}_i) \hat{X}(\overline{\lambda}^{\mathfrak{s}}_i) \overline{\phi}^{\mathfrak{s}}_i - h(\lambda^{\mathfrak{s}}_i) \hat{X}(\lambda^{\mathfrak{s}}_i) \phi^{\mathfrak{s}}_i \bigg\Vert_2
    \\
    &= \bigg\Vert \sum_i h(\overline{\lambda}^{\mathfrak{s}}_i) \hat{X}(\overline{\lambda}^{\mathfrak{s}}_i) \overline{\phi}^{\mathfrak{s}}_i - h(\lambda^{\mathfrak{s}}_i) \hat{X}(\overline{\lambda}^{\mathfrak{s}}_i) \overline{\phi}^{\mathfrak{s}}_i 
    \\
    &\quad + h(\lambda^{\mathfrak{s}}_i) \hat{X}(\overline{\lambda}^{\mathfrak{s}}_i) \overline{\phi}^{\mathfrak{s}}_i - h(\lambda^{\mathfrak{s}}_i) \hat{X}(\lambda^{\mathfrak{s}}_i) \phi^{\mathfrak{s}}_i \bigg\Vert_2
    \\
    &\leq \bigg\Vert \sum_i \left( h(\overline{\lambda}^{\mathfrak{s}}_i) - h(\lambda^{\mathfrak{s}}_i) \right) \hat{X}( \overline{\lambda}^{\mathfrak{s}}_i) \overline{\phi}^{\mathfrak{s}}_i \bigg\Vert_2 \quad \textbf{(1)}
    \\
    &\quad + \bigg\Vert \sum_i h(\lambda^{\mathfrak{s}}_i) \left( \hat{X}(\overline{\lambda}^{\mathfrak{s}}_i) \overline{\phi}^{\mathfrak{s}}_i - \hat{X}(\lambda^{\mathfrak{s}}_i) \phi^{\mathfrak{s}}_i \right) \bigg\Vert_2 \quad \textbf{(2)}.
\end{aligned}
\end{equation}
For \textbf{(1)}, $\{ \overline{\phi}^{\mathfrak{s}}_i \}$ are normalized eigenfunctions, applying assumption about filters we have:
\begin{equation}
    \textbf{(1)} \leq A_h \frac{\Vert W^{\mathfrak{s}}_N - W^{\mathfrak{s}}_{t_N} \Vert_2}{(2\epsilon(N))^{\frac{1}{4}}}  \bigg\Vert\sum_i \hat{X}( \overline{\lambda}^{\mathfrak{s}}_i) \overline{\phi}^{\mathfrak{s}}_i\bigg\Vert_2,
\end{equation}
that is:
\begin{equation}
    \textbf{(1)} \leq A_h \frac{\Vert W^{\mathfrak{s}}_N - W^{\mathfrak{s}}_{t_N} \Vert_2}{(2\epsilon(N))^{\frac{1}{4}}} ||X^{\mathfrak{s}}||_2.
\end{equation}
For \textbf{(2)}, as $X=\sum_i \hat{X}(\overline{\lambda}^{\mathfrak{s}}_i) \overline{\phi}^{\mathfrak{s}}_i = \sum_i \hat{X}(\lambda^{\mathfrak{s}}_i) \phi^{\mathfrak{s}}_i$, then we have:
\begin{equation}
    \sum_i k \left( \hat{X}(\overline{\lambda}^{\mathfrak{s}}_i) \overline{\phi}^{\mathfrak{s}}_i - \hat{X}(\lambda^{\mathfrak{s}}_i) \phi^{\mathfrak{s}}_i \right) = 0 \quad k\in \mathbb{R},
\end{equation}
substituting (42) into \textbf{(2)}, we get:
\begin{equation}
\begin{aligned}
    \textbf{(2)} &= \bigg\Vert \sum_i  \left( h(\lambda^{\mathfrak{s}}_i) - k \right)  \left( \hat{X}(\overline{\lambda}^{\mathfrak{s}}_i) \overline{\phi}^{\mathfrak{s}}_i - \hat{X}(\lambda^{\mathfrak{s}}_i) \phi^{\mathfrak{s}}_i \right) \bigg\Vert_2
    \\
    &\leq 2 \Delta h(\lambda^{\mathfrak{s}}) ||X^{\mathfrak{s}}||_2,
\end{aligned}
\end{equation}
where $\Delta h(\lambda^{\mathfrak{s}}) = \mathop{\min}_{k \in \mathbb{R}} \mathop{\max}_{ \lambda^{\mathfrak{s}}_i}\left\{ |{h}(\lambda^{\mathfrak{s}}_i ) - k| \right\}$ for all convolutional filters ${h}$, and $\{\lambda^{\mathfrak{s}}_i\}_{i \in \mathbb{Z} \setminus \{0\} }$ are eigenvalues of the stretched graphon $W^{\mathfrak{s}}_{t_N}$.
Combining the above inequalities, we get:
\begin{equation}
\begin{aligned}
    &|| T_{\overline{W}^{\mathfrak{s}}_N} \overline{X}^{\mathfrak{s}}_N - T_{{W}^{\mathfrak{s}}_{t_N}} X^{\mathfrak{s}} ||_2
    \\
    &\leq A_h \frac{\Vert W^{\mathfrak{s}}_N - W^{\mathfrak{s}}_{t_N} \Vert_2}{(2\epsilon(N))^{\frac{1}{4}}} ||X^{\mathfrak{s}}||_2 + 2 \Delta h(\lambda^{\mathfrak{s}}) ||X^{\mathfrak{s}}||_2
    \\
    &\quad + ||\overline{X}^{\mathfrak{s}}_N - X^{\mathfrak{s}}||_2.
\end{aligned}
\end{equation}
Similarly, we have:
\begin{equation}
\begin{aligned}
    &|| T_{\overline{W}^{\mathfrak{s}}_n} \overline{X}^{\mathfrak{s}}_n - T_{{W}^{\mathfrak{s}}_{t_N}} X^{\mathfrak{s}} ||_2
    \\
    &\leq A_h \frac{\Vert W^{\mathfrak{s}}_n - W^{\mathfrak{s}}_{t_N} \Vert_2}{(2\epsilon(n))^{\frac{1}{4}}} ||X^{\mathfrak{s}}||_2 + 2 \Delta h(\lambda^{\mathfrak{s}}) ||X^{\mathfrak{s}}||_2
    \\
    &\quad + ||\overline{X}^{\mathfrak{s}}_n - X^{\mathfrak{s}}||_2,
\end{aligned}
\end{equation}
therefore, we get the conclusion of the theorem:
\begin{equation}
\begin{aligned}
    &\mathbb{E}( ||\xi^{\mathfrak{s}} ( \boldsymbol{y}_N ) - \xi^{\mathfrak{s}} ( \boldsymbol{y}_n )||_2 )
    \\
    &\leq \sqrt{2}A_h ||X^{\mathfrak{s}}||_2 \left( \sqrt{\frac{L_1 - L_2^2}{L_1}} + \frac{A_{\mathbb{R}_+}t_N}{\sqrt{6d(N)}} \left( 1 + \frac{N}{n} \right) \right)
    \\
    &\quad + \frac{A_s}{\sqrt{6\sqrt{2d(N)}} } \left( \frac{1}{N^{\frac{3}{4}}} + \frac{ N^{\frac{1}{4}} }{\sqrt{n}} \right) + 4\Delta h(\lambda^{\mathfrak{s}}) ||X^{\mathfrak{s}}||_2.
\end{aligned}
\end{equation}

\section{Proof of Theorem \ref{theorem_of_downsampling}}
In this section, we first derive an expected upper bound for the transfer error of \textbf{sparse models without restrictions on sparsity}. Based on this, we obtain the conclusion of Theorem \ref{theorem_of_downsampling}. Both of these conclusions share the same increasing and decreasing trends concerning $N$, $n$, and $d$.

For $\mathbb{E} \left\{ \left\Vert \xi^{\mathfrak{s}} \left( \Phi(\widetilde{\mathbf{S}}_N, \boldsymbol{x}_N, \boldsymbol{\mathcal{H}}) \right) - \xi^{\mathfrak{s}} \left( \Phi(\widetilde{\mathbf{S}}_n, \boldsymbol{x}_n, \boldsymbol{\mathcal{H}}) \right) \right\Vert_2 \right\}$, we also transform the graphs and graph signals into their continuous forms. For simplicity of expression, here we still let $(\overline{W}^{\mathfrak{s}}_N, \overline{X}^{\mathfrak{s}}_N)$ denote the continuous forms of the large-scale graph and graph signal $(\widetilde{{\mathbf{S}}}_N, \boldsymbol{x}_N)$ obtained from the sparse random graph model, let $(\overline{W}^{\mathfrak{s}}_n, \overline{X}^{\mathfrak{s}}_n)$ denote the continuous forms of the smaller-scale graph and graph signal $(\widetilde{{\mathbf{S}}}_n, \boldsymbol{x}_n)$ obtained from the downsampling method. Then we have:
\begin{equation}
\begin{aligned}
    Y_N &= \xi^{\mathfrak{s}} \left( \Phi(\widetilde{\mathbf{S}}_N, \boldsymbol{x}_N, \boldsymbol{\mathcal{H}}) \right) = \Phi(\overline{W}^{\mathfrak{s}}_N, \overline{X}^{\mathfrak{s}}_N, \boldsymbol{\mathcal{H}})
    \\
    Y_n &= \xi^{\mathfrak{s}} \left( \Phi(\widetilde{\mathbf{S}}_n, \boldsymbol{x}_n, \boldsymbol{\mathcal{H}}) \right) = \Phi(\overline{W}^{\mathfrak{s}}_n, \overline{X}^{\mathfrak{s}}_n, \boldsymbol{\mathcal{H}}).
\end{aligned}
\end{equation}

In practical applications, GCNs often use a normalized adjacency matrix. Therefore, in our analysis, we also consider the normalization of the adjacency matrix, set $\widetilde{\mathbf{S}}_n =  \mathbf{S}_n / (\sqrt{ 2\epsilon(n)} n )$, since in a sparse graph model, sparsity is reflected in edge density, which is related to the expected average degree of the graph. 

To analysis the difference between the outputs of convolutional networks, we start from the last layer's output features and denote the output of $l$th layer by $X_{f_l}^{\{N\}}$ and $X_{f_l}^{\{n\}}$:
\begin{equation}
    ||Y_N - Y_n||_2^2 = \sum_{f_L=1}^{F_L} \left\Vert X_{f_L}^{\{N\}} - X_{f_L}^{\{n\}}\right\Vert_2^2.
\end{equation}
As the aggregation of WNN's layers can be represented as convolutional filters:
\begin{equation}
    X_{f_L}^{\{N\}} = \sigma \left( \sum_{f_{L-1}}^{F_{L-1}} h_{f_{L-1},f_L}(\overline{W}^{\mathfrak{s}}_N) * X_{f_{L-1}}^{\{N\}} \right),
\end{equation}
for simplicity of expression, we use $h_L$ instead of $h_{f_{L-1},f_L}$ hereafter. The activation functions are normalized Lipschitz, we derive the above difference equation into:
\begin{equation}
\begin{aligned}
    &\left\Vert X_{f_L}^{\{N\}} - X_{f_L}^{\{n\}} \right\Vert_2 
    \\
    &\leq \left\Vert  \sum_{f_{L-1}}^{F_{L-1}} h_L(\overline{W}^{\mathfrak{s}}_N) * X_{f_{L-1}}^{\{N\}}  
    -  h_L(\overline{W}^{\mathfrak{s}}_n) * X_{f_{L-1}}^{\{n\}}  \right\Vert_2
    \\
    &\leq \sum_{f_{L-1}}^{F_{L-1}} \left\Vert h_L(\overline{W}^{\mathfrak{s}}_N) * X_{f_{L-1}}^{\{N\}}  
    -  h_L(\overline{W}^{\mathfrak{s}}_n) * X_{f_{L-1}}^{\{n\}}  \right\Vert_2.
\end{aligned}
\end{equation}
Assume there exists a series of intermediate variables related to SWNN $\Phi(W^{\mathfrak{s}}_{t_N}, X^{\mathfrak{s}}, \boldsymbol{\mathcal{H}}), W^{\mathfrak{s}}_{t_N} = W_{t_N} / \sqrt{2\epsilon(N)}$, and the output of $l$th layer is $X^{\mathfrak{s}}_{f_l}$. Using triangle equality, we get:
\begin{equation}
\begin{aligned}
    &\left\Vert h_L(\overline{W}^{\mathfrak{s}}_N) * X_{f_{L-1}}^{\{N\}}  
    -  h_L(\overline{W}^{\mathfrak{s}}_n) * X_{f_{L-1}}^{\{n\}}  \right\Vert_2
    \\
    &\leq \left\Vert h_L(\overline{W}^{\mathfrak{s}}_N) * X_{f_{L-1}}^{\{N\}}  
    -  h_L({W}^{\mathfrak{s}}_{t_N}) * X^{\mathfrak{s}}_{f_{L-1}}  \right\Vert_2
    \\
    &\quad +  \left\Vert h_L({W}^{\mathfrak{s}}_{t_N}) * X^{\mathfrak{s}}_{f_{L-1}}  
    -  h_L(\overline{W}^{\mathfrak{s}}_n) * X_{f_{L-1}}^{\{n\}}  \right\Vert_2,
\end{aligned}
\end{equation}
similarly to the proof of Theorem \ref{theorem_con}, we have:
\begin{equation}
\begin{aligned}
    &\left\Vert X_{f_L}^{\{N\}} - X_{f_L}^{\{n\}} \right\Vert_2
    \\
    &\leq \sum_{f_{L-1}}^{F_{L-1}} {A_h \left\Vert X^{\mathfrak{s}}_{f_{L-1}} \right\Vert_2} \left( \frac{\Vert W^{\mathfrak{s}}_N - W^{\mathfrak{s}}_{t_N} \Vert_2}{(2\epsilon(N))^{\frac{1}{4}}} + \frac{\Vert W^{\mathfrak{s}}_n - W^{\mathfrak{s}}_{t_N} \Vert_2}{(2\epsilon(n))^{\frac{1}{4}}} \right) 
    \\
    &\quad + \sum_{f_{L-1}}^{F_{L-1}} \left( \left\Vert X_{f_{L-1}}^{\{N\}} - X^{\mathfrak{s}}_{f_{L-1}} \right\Vert_2 + \left\Vert X_{f_{L-1}}^{\{n\}} - X^{\mathfrak{s}}_{f_{L-1}} \right\Vert_2\right)
    \\
    &\quad + \sum_{f_{L-1}}^{F_{L-1}} 4 \left\Vert X^{\mathfrak{s}}_{f_{L-1}} \right\Vert_2 \Delta h(\lambda^{\mathfrak{s}}),
\label{kernel_of_trans_theorem}
\end{aligned}
\end{equation}
where $\Delta h(\lambda^{\mathfrak{s}}) = \mathop{\min}_{k \in \mathbb{R}} \mathop{\max}_{ \lambda^{\mathfrak{s}}_i}\left\{ |{h}(\lambda^{\mathfrak{s}}_i ) - k| \right\}$ for all convolutional filters ${h}$, and $\{\lambda^{\mathfrak{s}}_i, i \in \mathbb{Z} \setminus \{0\} \}$ are eigenvalues of the graphon $W^{\mathfrak{s}}_{t_N}$.

Using the assumption about activation functions and $\sigma(0)=0$, that is $ |\sigma(x)-\sigma(0)| \leq |x| $, then $||X^{\mathfrak{s}}_{f_{L-1}}||_2$ can be written as:
\begin{equation}
    ||X^{\mathfrak{s}}_{f_{L-1}}||_2 \leq \left\Vert \sum_{f_{L-2}}^{F_{L-2}} h_L({W}^{\mathfrak{s}}_{t_N}) * X^{\mathfrak{s}}_{f_{L-2}} \right\Vert_2,
\end{equation}
considering filters are non-amplifying and using Cauchy Schwarz inequalities, we get:
\begin{equation}
\begin{aligned}
    ||X^{\mathfrak{s}}_{f_{L-1}}||_2 
    &\leq \sum_{f_{L-2}}^{F_{L-2}} \left\Vert  h_L({W}^{\mathfrak{s}}_{t_N}) * X^{\mathfrak{s}}_{f_{L-2}} \right\Vert_2
    \\
    &\leq \sum_{f_{L-2}}^{F_{L-2}} \left\Vert  X_{f^{\mathfrak{s}}_{L-2}} \right\Vert_2
    \\
    &\leq \prod_{l=1}^{L-2} F_l \sum_{f_0}^{F_0} ||X^{\mathfrak{s}}_{f_0}||_2.
\label{recur_x_L}
\end{aligned}
\end{equation}
Expanding (\ref{kernel_of_trans_theorem}) recursively, and substituting the results of (\ref{recur_x_L}), we have:
\begin{equation}
\begin{aligned}
    &\left\Vert X_{f_L}^{\{N\}} - X_{f_L}^{\{n\}} \right\Vert_2
    \\
    &\leq L \prod_{l=1}^{L-1} F_l \sum_{f_0}^{F_0} \left\Vert X^{\mathfrak{s}}_{f_0} \right\Vert_2 {A_h }  \bigg( \frac{\Vert W^{\mathfrak{s}}_N - W^{\mathfrak{s}}_{t_N} \Vert_2}{(2\epsilon(N))^{\frac{1}{4}}} + 4\Delta h(\lambda)
    \\
    &\quad + \frac{\Vert W^{\mathfrak{s}}_n - W^{\mathfrak{s}}_{t_N} \Vert_2}{(2\epsilon(n))^{\frac{1}{4}}}  \bigg) 
    \\
    &\quad + F_0 \left( \left\Vert X_{f_0}^{\{N\}} - X^{\mathfrak{s}}_{f_0} \right\Vert_2 + \left\Vert X_{f_{0}}^{\{n\}} - X^{\mathfrak{s}}_{f_{0}} \right\Vert_2\right),
\label{trans_theorem_last_2}
\end{aligned}
\end{equation}
where $X_{f_0}^{\{N\}} = \overline{X}^{\mathfrak{s}}_N, X_{f_0}^{\{n\}} = \overline{X}^{\mathfrak{s}}_n, X_{f_0}=X$. Since $F_0 = F_L = 1$ and $F_l=F$ for $1 \leq l \leq L-1$, we have $Y_N = X_{f_L}^{\{N\}}$ and $Y_n = X_{f_L}^{\{n\}}$. From the conclusions about sampling expectation lemmas we have:
\begin{equation}
    \begin{aligned}
        \mathbb{E} \left\{ \left\Vert \overline{X}^{\mathfrak{s}}_N - X^{\mathfrak{s}} \right\Vert_2 \right\} &\leq \frac{A_s}{\sqrt{6N \sqrt{2\epsilon(N)}}}
        \\
        \mathbb{E} \left\{ \left\Vert \overline{X}^{\mathfrak{s}}_n - X^{\mathfrak{s}} \right\Vert_2 \right\} &\leq \frac{A_s}{\sqrt{6n \sqrt{2\epsilon(n)}}}
        \\
        \mathbb{E} \left\{ \left\Vert \overline{W}^{\mathfrak{s}}_N - {W}^{\mathfrak{s}}_{t_N} \right\Vert_2 \right\} &\leq  \frac{1}{(2\epsilon(N))^{\frac{1}{4}}} \left( \frac{\sqrt{L_1 - L_2^2}}{t_N} + \frac{2A_{\mathbb{R}_+}t_N}{\sqrt{6N}} \right)
        \\
        \mathbb{E} \left\{ \left\Vert \overline{W}^{\mathfrak{s}}_n - {W}^{\mathfrak{s}}_{t_N} \right\Vert_2 \right\} &\leq \frac{1}{(2\epsilon(n))^{\frac{1}{4}}} \left( \frac{\sqrt{L_1 - L_2^2}}{t_N} + \frac{2A_{\mathbb{R}_+}t_N}{\sqrt{6n}} \right),
    \end{aligned}
\end{equation}
substituting these equations and inequalities into (\ref{trans_theorem_last_2}), then we get:
\begin{equation}
\begin{aligned}
    &\mathbb{E} \left\{\Vert Y_N - Y_n \Vert_2 \right\}
    \\
    &\leq 2 C_1 {A_h } \frac{1}{ \sqrt{2\epsilon(N)} } \bigg( \frac{\sqrt{L_1 - L_2^2}}{t_N} + \frac{A_{\mathbb{R}_+}t_N}{\sqrt{6N}}  + \frac{A_{\mathbb{R}_+}t_N}{\sqrt{6n}} \bigg) 
    \\
    &\quad + \frac{1}{(2\epsilon(N))^{\frac{1}{4}}} \left( \frac{A_s}{\sqrt{6N}} + \frac{A_s}{\sqrt{6n}} \right) + 4 C_1 \Delta h(\lambda),
\label{trans_theorem_last_1}
\end{aligned}
\end{equation}
where $C_1 = LF^{L-1}||X^{\mathfrak{s}}||_2$.
For the topological structures of the original large-scale graph, we have:
\begin{equation}
    \begin{aligned}
        \epsilon(N) = \frac{L_1}{t_N^2},\quad 
        d(N) \approx N\epsilon(N),
    \end{aligned}
\end{equation}
therefore, the first part of the right side in (\ref{trans_theorem_last_1}) can be written as:
\begin{equation}
    \begin{aligned}
        &\frac{1}{ \sqrt{\epsilon(N)} } \bigg( \frac{\sqrt{L_1 - L_2^2}}{t_N} + \frac{A_{\mathbb{R}_+}t_N}{\sqrt{6N}}  + \frac{A_{\mathbb{R}_+}t_N}{\sqrt{6n}} \bigg)
        \\
        &= \sqrt{ 1 - \frac{L_2^2}{L_1} }  + \frac{A_{\mathbb{R}_+} t_N}{ \sqrt{6N\epsilon(N)}} \bigg( 1 + \sqrt{\frac{N}{n}} \bigg)
        \\
        &= \sqrt{ 1 - \frac{L_2^2}{L_1} }  + \frac{A_{\mathbb{R}_+}}{ \sqrt{6}} \frac{t_N }{ \sqrt{d} } \bigg( 1 + \sqrt{\frac{N}{n}} \bigg),
    \end{aligned}
\end{equation} 
substituting the above equation back to (\ref{trans_theorem_last_1}), then we have:
\begin{equation}
\begin{aligned}
    &\mathbb{E} \left\{\Vert Y_N - Y_n \Vert_2 \right\}
    \\
    &\leq \sqrt{2} C_1 A_h \bigg( \sqrt{ 1 - \frac{L_2^2}{L_1} }  + \frac{A_{\mathbb{R}_+}}{ \sqrt{6}} \frac{t_N }{ \sqrt{d} } \bigg( 1 + \sqrt{\frac{N}{n}} \bigg) \bigg)
    \\
    &\quad + \frac{N^{1/4}}{\sqrt{6\sqrt{2d(N)}}} \left( \frac{1}{\sqrt{N}} + \frac{1}{\sqrt{n}} \right) + 4 C_1 \Delta h(\lambda^{\mathfrak{s}}),
\label{theorem_all_sparsity}
\end{aligned}
\end{equation}
it's the expected upper bound for the transfer error of \textbf{sparse models without restrictions on sparsity},
where $L_1 = ||W_{\mathbb{R}_+}||_{L_2}$ in $[0,t_N]^2$ and $L_2 = ||W_{\mathbb{R}_+}||_{L_2}$ in $[0,t_N]^2$ make $\sqrt{ 1 - \frac{L_2^2}{L_1} }$ decrease about $d(N)$ and increase about $N$, and $t_N$ increases with $N$.

Considering the sparse model with $t_N=\sqrt{N}$ in theorem \ref{theorem_of_downsampling}, and $\sqrt{1-\frac{L^2_2}{L_1}} \leq 1$, then we prove theorem \ref{theorem_of_downsampling}:
\begin{equation}
\begin{aligned}
    &\mathbb{E} \left\{\Vert Y_N - Y_n \Vert_2 \right\}
    \\
    &\leq \sqrt{2}C_1 A_h \bigg(  1 + \frac{A_{\mathbb{R}_+}}{ \sqrt{6}} \sqrt{\frac{N}{d}} \bigg( 1 + \sqrt{\frac{N}{n}} \bigg) \bigg)
    \\
    &\quad + \frac{N^{1/4}}{\sqrt{6\sqrt{2d}}} \left( \frac{1}{\sqrt{N}} + \frac{1}{\sqrt{n}} \right) + 4 C_1 \Delta h(\lambda),
\label{theorem_specific}
\end{aligned}
\end{equation}
both of these conclusions share the same increasing and decreasing trends concerning $N$, $n$, and $d$.

\bibliographystyle{IEEEtran}
\bibliography{IEEEabrv, references}

\end{document}